\documentclass[preprint,12pt]{elsarticle}
\usepackage{booktabs}
\usepackage{subcaption}
\usepackage[colorlinks=true, linkcolor=blue, citecolor=blue, urlcolor=blue]{hyperref}
\hypersetup{colorlinks=true}

\usepackage{amssymb}
\usepackage{amsmath}

\journal{Computers in Biology and Medicine}

\begin{document}

\begin{frontmatter}



\title{Alzheimer’s Disease Prediction Using EffNetViTLoRA and BiLSTM with Multimodal Longitudinal MRI Data}

\author[1]{Mahdieh Behjat Khatooni}
\ead{m_behjat@comp.iust.ac.ir}

\author[1]{Mohsen Soryani\corref{cor1}}

\cortext[cor1]{Corresponding author:
	Tel.: +982173225304;
	Fax: +982173021220;
	Email: soryani@iust.ac.ir}

\affiliation[1]{
	organization={School of Computer Engineering, Iran University of Science and Technology},
	city={Tehran},
	country={Iran}
}

\begin{abstract}
Alzheimer’s disease (AD) is a prevalent neurodegenerative disorder that progressively impairs memory, decision-making, and overall cognitive function. As AD is irreversible, early prediction is critical for timely intervention and management. Mild Cognitive Impairment (MCI), a transitional stage between cognitively normal (CN) aging and AD, plays a significant role in early AD diagnosis. However, predicting MCI progression remains a significant challenge, as not all individuals with MCI convert to AD. MCI subjects are categorized into stable MCI (sMCI) and progressive MCI (pMCI) based on conversion status. In this study, we propose a generalized, end-to-end deep learning model for AD prediction using MCI cases from the Alzheimer’s Disease Neuroimaging Initiative (ADNI). Our hybrid architecture integrates Convolutional Neural Networks and Vision Transformers to capture both local spatial features and global contextual dependencies from Magnetic Resonance Imaging (MRI) scans. To incorporate temporal progression, we further employ Bidirectional Long Short-Term Memory (BiLSTM) networks to process features extracted from four consecutive MRI timepoints along with some other non-image biomarkers, predicting each subject’s cognitive status at month 48. Our multimodal model achieved an average progression prediction accuracy of 95.05\% between sMCI and pMCI, outperforming existing studies in AD prediction. This work demonstrates state-of-the-art performance in longitudinal AD prediction and highlights the effectiveness of combining spatial and temporal modeling for the early detection of Alzheimer’s disease.
\end{abstract}



\end{frontmatter}

\section{Introduction}
\label{sec:introduction}
Neurodegenerative disorders are diseases that progressively affect the life of neurons in the brain. Alzheimer’s disease (AD) is the most prevalent neurodegenerative disorder worldwide. It is the leading cause of dementia, progressive and irreversible, with no definite cure at present. Neurons are responsible for transmitting messages between different parts of the brain and from the brain to the muscles and organs. In AD, neurons in specific brain regions are destroyed, leading to significant disruptions in cognitive and perceptual functions. The disease primarily affects memory, thinking, and behavior due to progressive brain cell loss. Neuronal death in memory-related regions, particularly the medial temporal lobe including the hippocampus \cite{borbely2013neuropeptides}, is a well-known hallmark of AD. As the disease progresses, cognitive decline intensifies, initially manifesting as forgetfulness. Over time, individuals lose the ability to perform even simple tasks, ultimately becoming entirely dependent on others for basic daily activities \cite{AlzheimerSymptoms}. The affected population is expected to rise to approximately 152 million by 2050. The current global cost of the disease is around one trillion US dollars annually and is projected to double by 2030 \cite{patterson2018world}. Given its irreversible nature, early prediction of Alzheimer’s disease is essential.
In recent decades, artificial intelligence and machine learning, particularly deep learning \cite{lecun2015deep}, has provided promising tools for understanding and identifying connections within different regions of brain images. These advancements have significantly contributed to the diagnosis and prognosis of Alzheimer’s disease, supporting the medical community by analyzing structural and functional patterns in brain images obtained from various imaging modalities, as well as non-imaging data. Deep learning methods excel at analyzing complex, high-dimensional datasets and uncovering hidden patterns and features, making them indispensable for addressing the challenges of Alzheimer’s disease diagnosis and prognosis.

Alzheimer’s disease is typically categorized into three stages based on progression:
\begin{enumerate}
	\item Cognitively Normal (CN): Individuals with no signs of cognitive impairment.
	\item Mild Cognitive Impairment (MCI): An intermediate stage between CN and AD that is often targeted for early intervention to prevent progression to AD.
	\item Alzheimer’s Disease (AD): Patients who exhibit full clinical symptoms of Alzheimer’s disease.
\end{enumerate}	
MCI is further categorized into two subgroups based on progression. Stable MCI (sMCI), also known as non-converter MCI in the literature, refers to cases that remain stable and do not progress to AD over time. Progressive MCI (pMCI), also known as converter MCI, refers to cases that progress to AD within a follow-up period. Early and accurate prediction of MCI progression, stable MCI versus progressive MCI, is crucial as it helps physicians monitor and potentially slow the progression of the disease.

Multiple imaging modalities are commonly used for the diagnosis and prediction of Alzheimer’s disease in computer-aided diagnosis (CAD) systems based on artificial intelligence, including Positron Emission Tomography (PET), functional Magnetic Resonance Imaging, and structural Magnetic Resonance Imaging. Structural Magnetic Resonance Imaging (sMRI) is a widely used modality in AD prediction research, because it is non-invasive and capable of capturing high-resolution images of subtle brain structural changes associated with the disease.

Recent advances in deep learning, especially using Convolutional Neural Networks (CNNs) and Vision Transformers (ViTs) \cite{dosovitskiy2020image}, have improved the automated analysis of medical images. Given that AD is a slowly progressing condition that can take years to develop from a normal cognitive state, tracking its progression over time is vital. A high percentage of recent studies on AD prediction using machine learning, which propose CAD systems, have a major limitation. They rely on single time-point image data to assess the current state of the patient. In contrast, longitudinal imaging data are highly valuable for capturing temporal changes in the brain and identifying patterns that could predict AD before it fully manifests. This area holds great potential and warrants further research to support early prediction and timely intervention. Recurrent Neural Networks (RNNs) and their advanced variants, Long Short-Term Memory (LSTM) and Bidirectional Long Short-Term Memory (BiLSTM), have proven effective in modeling temporal dependencies in longitudinal data, making them particularly well-suited for analyzing time-series medical data.

In this paper, we used sequential MRI follow-up data as well as non-imaging biomarkers to predict AD within four years. We review recent research on Alzheimer’s disease prediction using follow-up data in Section \ref{sec:related-works}. Section \ref{sec:methodology} provides a detailed description of the proposed prediction method, presented in two phases: feature extraction and prediction. In Section \ref{sec:result}, we introduce the dataset and preprocessing steps, followed by the results of the prediction model with interpretation. To further validate the proposed method, we conduct an ablation study that analyzes the contribution of individual components to overall performance. Finally, Section \ref{sec:conclusion} summarizes the key findings and concludes the study.

\section{Related Work}
\label{sec:related-works}
Machine learning and deep learning have played a central role in recent research on medical image analysis, providing powerful tools for feature extraction, pattern recognition, and disease prediction. While many earlier studies in the context of Alzheimer’s disease diagnosis and prediction focused on single time-point imaging, recent research has emphasized the importance of longitudinal imaging data, which capture disease progression over time and provide richer information for prediction. In this section, we review recent studies that have used longitudinal imaging data to predict Alzheimer’s disease.

In \cite{zhu2021long}, authors proposed an innovative temporal-structural Support Vector Machine (TS-SVM) framework for early detection of MCI-to-AD conversion using longitudinal MRI data. The method requires only two consecutive scans and is capable of predicting AD up to 12 months before its clinical diagnosis. The model preserves temporal consistency by enforcing monotonic risk scores that increase over time, eliminating unreasonable fluctuations. It further incorporates a joint feature selection and classification stage to extract the most discriminative morphological features. This approach achieved an accuracy of 81.75\% in predicting AD progression. In \cite{er2020predicting}, a hybrid framework utilizing voxel-based morphometry (VBM), autoencoders, CNNs, and SVMs was proposed to predict Alzheimer’s conversion in MCI patients using longitudinal 3D MRI data. The model achieved strong classification results by focusing on prognostic features from anatomically relevant regions. Authors in \cite{abuhmed2021robust} proposed a hybrid multimodal multitask deep learning framework using BiLSTM to process multivariate time-series data across four visits, alongside baseline static features. Temporal and static features are fused to support two strategies: direct classification and multitask regression-based learning, where predicted cognitive scores are used for final progression classification. In \cite{hoang2023vision}, authors employed the original Vision Transformer architecture pretrained on ImageNet \cite{deng2009imagenet} for Alzheimer’s disease prediction based on baseline visits only. VGG-TSwinformer, a hybrid deep learning model that combines VGG-16 \cite{simonyan2014very} and a temporal–spatial transformer with sliding-window attention \cite{vaswani2017attention}, was proposed to model brain atrophy progression in MCI patients using longitudinal sMRI scans taken two years apart \cite{hu2023vgg}. Performing binary classification of sMCI versus pMCI, their model achieved 77.2\% accuracy and an AUC of 81.53\%, outperforming cross-sectional MRI approaches. A model based on LSTM was proposed in \cite{el2022two} that leverages multimodal data, including hand-crafted MRI features and cognitive scores, to predict Alzheimer’s disease up to 48 months in advance using longitudinal data. Additionally, the model estimates the time of conversion for progressive cases by predicting the number of months until conversion. The study demonstrated that integrating discriminative feature selection with an LSTM classifier enabled three-class prediction of AD progression with an average accuracy of 93.87\% on the validation set and 91.22\% on the test set. In \cite{el2020multimodal}, a multimodal multitask model combining CNNs and BiLSTMs leverages longitudinal data to jointly classify Alzheimer’s progression and predict cognitive scores. By modeling temporal dynamics across 15 time points, the method achieved over 92\% accuracy and 98\% recall in the four-class classification of AD, sMCI, pMCI, and CN, highlighting the value of multimodal longitudinal data. However, the framework requires external feature extraction and is not fully end-to-end, and it includes AD and CN groups whose transitions are not meaningful, as they typically remain in the same group over time in AD prediction. A deep learning approach for predicting Alzheimer’s disease progression using longitudinal MRI and neuropsychological data was presented in \cite{lei2020deep}. The framework applies LASSO\footnote{Least Absolute Shrinkage and Selection Operator} for feature selection, imputes missing values from earlier visits, and uses Correntropy to mitigate noise, thereby capturing temporal patterns for improved prediction. In \cite{gao2023hybrid}, a hybrid model integrating convolutional neural networks with attention mechanisms and age information was proposed for Alzheimer’s disease diagnosis and prediction. A multi-scale attention-based convolutional subnetwork extracts rich features from 3D MRI images, while a separate age-transformer subnetwork encodes age information. These subnetworks are fused in an end-to-end framework to predict disease probabilities. The model achieves over 90\% accuracy in AD vs. CN classification and over 73\% accuracy in predicting sMCI vs. pMCI, leveraging pre-trained weights from AD/CN classification for the more challenging MCI prediction task.

\section{Method}
\label{sec:methodology}
\begin{figure}
	\centering
	\includegraphics[width=0.8\linewidth]{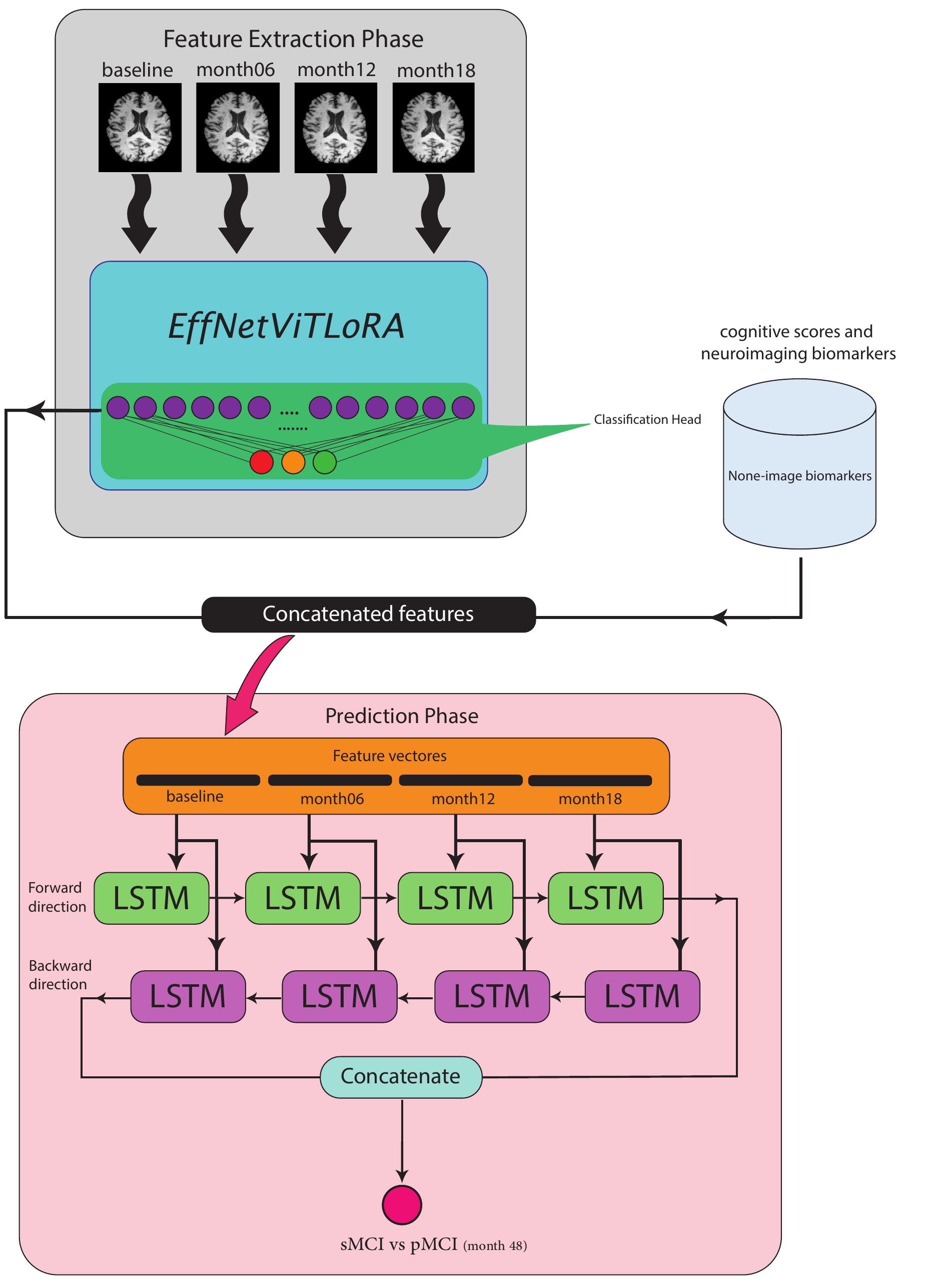}
	\caption{Diagram of the proposed prediction method, consisting of two phases. Phase 1 is the EffNetViTLoRA, which serves as a feature extractor. The representative features of the MCI group are obtained from the last hidden layer, and 17 additional non-image biomarkers are appended to each image feature vector to construct the multimodal prediction data. Phase 2 is the prediction phase, which utilizes the features obtained from Phase 1 in combination with the biomarkers to predict Alzheimer’s disease progression based on follow-up data four years after the baseline visit.}
	\label{fig:predictDiagram}
\end{figure}

The proposed prediction method comprises two separate phases. In the first phase, we utilized a hybrid diagnostic model originally introduced in \cite{khatooni2025effnetvitloraefficienthybriddeep}, consisting of a CNN and a ViT, which processes MRI data to classify each image into one of three categories. In the second phase, we used the abstract representative features extracted in the first phase, along with additional non-image biomarkers of the MCI group, to form the follow-up data used in predicting cognitive status at month 48. Fig. \ref{fig:predictDiagram} provides an overview of the proposed model, which consists of distinct feature extraction and prediction phases.

\subsection{Feature extraction model (EffNetViTLoRA)}
\begin{figure}
	\centering
	\includegraphics[width=\linewidth]{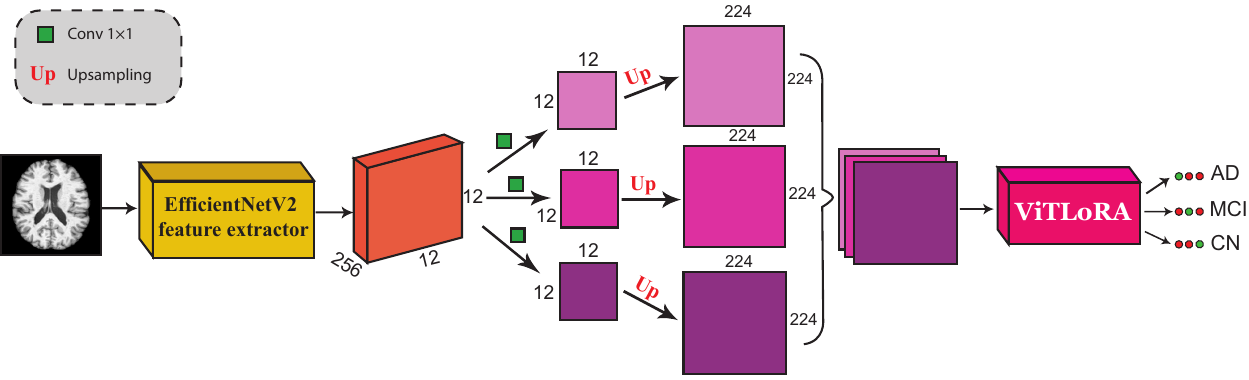}
	\caption{Architecture of the hybrid feature extraction model. The model captures both local and global dependencies in input MRI images by combining EfficientNetV2 with a pretrained ViT. The ViT is fine-tuned using LoRA applied to the Key, Query, and Value weight matrices.}
	\label{fig:hybrid}
\end{figure}

To effectively capture both local and global dependencies in MRI brain images, we employed a hybrid model named EffNetViTLoRA \cite{khatooni2025effnetvitloraefficienthybriddeep} combining CNN and ViT architectures. Fig. \ref{fig:hybrid} provides an abstract representation of the model architecture. First, processed MRI images are fed into an EfficientNetV2 Small \cite{tan2021efficientnetv2}, pretrained on ImageNet, with its final classification layer removed to extract local features. To further enhance feature extraction, the resulting feature maps are then passed to a pretrained ViT with 16×16 patches. To ensure efficient training and optimal performance without requiring extensive fine-tuning of a large model, we froze all layers of the ViT. Instead, we adapted it to our target dataset using Low-Rank Adaptation (LoRA) \cite{hu2021lora}, applied to the Key, Query, and Value weight matrices in each ViT encoder.
To match the output feature maps from EfficientNet to the ViT input size, we applied three 1×1 convolution layers followed by interpolation to achieve a final dimension of 3×256×256. The most representative features are extracted from the model’s final hidden layer, resulting in a 256-dimensional abstract feature vector. For each subject in our dataset, a sequence of four time points, each represented by a 256-dimensional feature vector, constitutes the final feature sequence.
\subsection{Prediction Model}
\label{sec:prediction_model}
\begin{figure}
	\centering
	\includegraphics[width=0.8\linewidth]{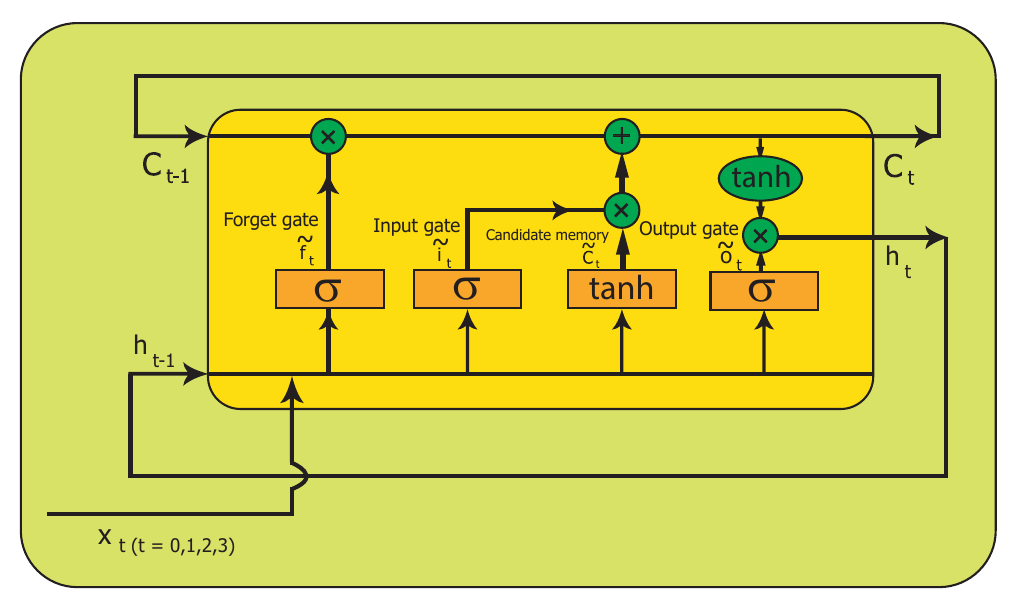}
	\caption{Internal structure of an LSTM cell.}
	\label{fig:LSTM_architecture}
\end{figure}

BiLSTM is a variant of LSTM networks which is an effective recurrent model for capturing temporal relationships in both forward and backward directions in time-series data. In this study, we designed a BiLSTM model with a single layer of 256 hidden units in both the forward and backward LSTMs to capture dependencies between feature tokens across time. The unfolded LSTM cell is shown in Figure~\ref{fig:LSTM_architecture}, where each unit consists of an input gate, a forget gate, and an output gate that regulate information flow across time steps. This structure enables the BiLSTM to capture long-term dependencies by processing sequences in both forward and backward directions. The internal operations of the LSTM cell are governed by mathematical equations as follows:
\begin{equation}
	f_t = \sigma(W_f [h_{t-1}, x_t] + b_f)
\end{equation}

\begin{equation}
	i_t = \sigma(W_i [h_{t-1}, x_t] + b_i)
\end{equation}

\begin{equation}
	\tilde{c}_t = \tanh(W_c [h_{t-1}, x_t] + b_c)
\end{equation}

\begin{equation}
	c_t = f_t \odot c_{t-1} + i_t \odot \tilde{c}_t
\end{equation}

\begin{equation}
	o_t = \sigma(W_o [h_{t-1}, x_t] + b_o)
\end{equation}

\begin{equation}
	h_t = o_t \odot \tanh(c_t)
\end{equation}
Where $\sigma$ denotes the sigmoid activation function, and $\odot$ represents element-wise multiplication. The forget gate ($f_t$) controls the retention of past information, the input gate ($i_t$) determines how much new information is stored in the cell state, and the output gate ($o_t$) regulates the hidden state passed to the next step. The cell state ($c_t$) is updated through a combination of the previous cell state and the candidate memory content, ensuring the model can preserve long-term dependencies.

For each MCI subject, MRI scans from four time points (baseline, month 6, month 12, and month 18 visits) were processed through the feature extraction model, operating in inference mode. From the final hidden layer of EffNetViTLoRA, we extracted 256-dimensional feature vectors at each time point. To enrich the feature representation and improve prediction performance, we incorporated non-imaging biomarkers alongside the MRI derived features. These biomarkers include cognitive test scores and neuroimaging features, such as the volumes of brain regions affected by Alzheimer’s disease, as well as results from psychological assessments. After evaluating multiple groups of candidate biomarkers, we identified and retained a subset of 17 effective biomarkers for integration with the MRI features.
These 17 non-imaging biomarkers were concatenated with the MRI features at each time point, resulting in a time-series sequence of four steps, each with a feature length of 273 (256 from MRI + 17 biomarkers). The BiLSTM model processes this sequence to capture temporal dependencies within sequences and predict whether each subject will convert to Alzheimer’s disease or remain stable within the four years following baseline, using information up to the 18-month visit.

\section{Experiments and results}
\label{sec:result}
\subsection{Dataset and preprocessing}
\begin{figure}
	\centering
	\includegraphics[width=0.8\linewidth]{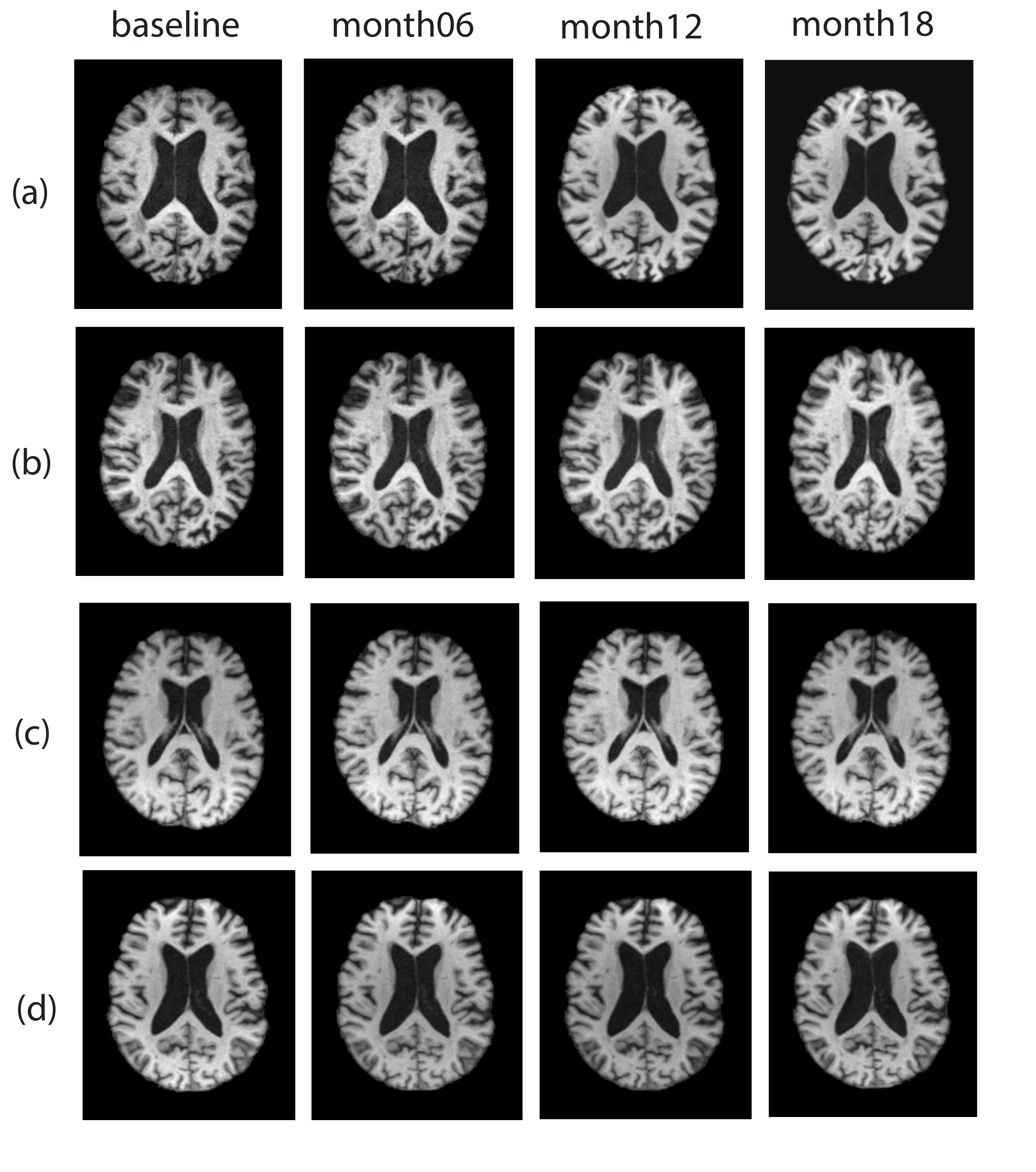}
	\caption{Sample axial slices from ADNI for sMCI and pMCI subjects over an 18-month follow-up. Rows (a) and (b) show sMCI cases, while rows (c) and (d) show pMCI cases. Subtle shrinkage in the gray matter and hippocampal regions can be observed across the four time points in the pMCI rows.}
	\label{fig:2classes}
\end{figure}

In this work, we used data from the Alzheimer’s Disease Neuroimaging Initiative (ADNI)\footnote{\url{http://adni.loni.usc.edu}}. ADNI was launched in 2003 as a public-private partnership, led by Principal Investigator Michael W. Weiner, MD. The primary goal of ADNI has been to test whether serial MRI, PET, other biological markers, and clinical and neuropsychological assessments can be combined to measure the progression of mild cognitive impairment and early Alzheimer’s disease. In the feature extraction phase, we utilized 2,010 subjects, using all T1-weighted MRI volumes available in the ADNI1, ADNIGO, ADNI2, and ADNI3 phases at the time of our experiments. While multiple MRI scans were available for most subjects, representing time-series data, we used only the baseline visit volumes of each subject for consistency. In the prediction phase, we used a total of 530 subjects from the sMCI and pMCI groups with follow-up data available for at least 48 months. Table \ref{tab:demographics} presents the detailed demographics and clinical information of the subjects included in both the feature extraction and prediction phases.
Two sets of sample slices for stable MCI and progressive MCI, representing subjects who remained stable at the MCI stage and those who converted to AD within 48 months, are presented in Fig.~\ref{fig:2classes}. On close inspection, no distinctive changes could be observed in the sMCI slices over time, whereas very subtle shrinkage in the gray matter and hippocampi could be tracked across the four consecutive time points in the pMCI slices. Atrophy in the gray matter and hippocampal regions of the medial temporal lobe is among the most noticeable effects of Alzheimer’s disease \cite{cavedo2014medial}.
\begin{table}
	\caption{Summary of the demographic information of the included subjects in the feature extraction and prediction phases.}
	\centering
	\label{tab:demographics}
	\begin{tabular}{@{}llll@{}}
		\toprule
		\textbf{Category}       & 
		\textbf{\# of subjects}    &
		\textbf{Gender (M/F)}    &
		\textbf{Age (years) Mean ± SD} \\ 
		\midrule
		\multicolumn{4}{l}{\textbf{Feature extraction phase}} \\ 
		\midrule
		AD             & 444      & 244/200         & 75.88 ± 07.91    \\
		MCI            & 706      & 415/291         & 76.07 ± 07.68    \\
		CN             & 860      & 375/485         & 76.28 ± 06.79    \\
		Combined       & 2010     & 1034/976        & 76.11 ± 07.36    \\
		\midrule
		\multicolumn{4}{l}{\textbf{Prediction phase}} \\ 
		\midrule
		sMCI           & 390      & 235/155         & 72.89 ± 07.59    \\
		pMCI           & 140      & 86/54           & 73.87 ± 06.97    \\
		\bottomrule
	\end{tabular}
\end{table}

\begin{figure}
	\centering
	\includegraphics[width=\linewidth]{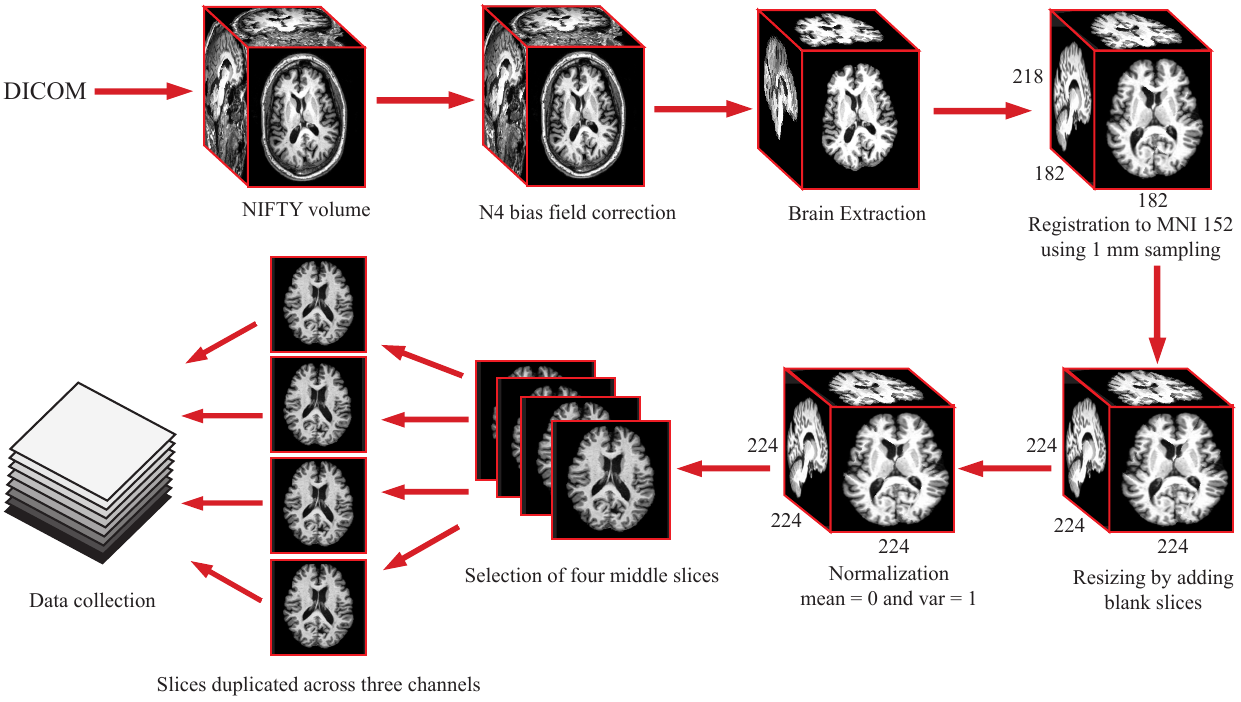}
	\caption{Preprocessing steps of ADNI MRI T1-weighted volumes in our study. The arrows demonstrate the flow of the process.}
	\label{fig:preprocess}
\end{figure}

Preprocessing steps are shown in Fig.~\ref{fig:preprocess}. DICOM layers were converted to NIfTI format using the MRICron software. To ensure consistent image quality, we applied N4 bias field correction \cite{tustison2010n4itk} to standardize image intensity. Next, the Brain Extraction Tool (BET) \cite{isensee2019automated} was used to remove non-brain tissue, isolating brain anatomy and enabling the model to focus exclusively on relevant structures. Each volume was spatially normalized to the MNI152 template space with a 1 mm sampling resolution using the FLIRT \cite{jenkinson2001global,jenkinson2002improved} registration tool in the FSL \cite{woolrich2009bayesian} software package, resulting in volumes of size 218×182×218.
To standardize dimensions for compatibility with the input requirements of the pretrained Vision Transformer, we added blank slices to the beginning and end of each volume, adjusting the dimensions to 224×224×224. Finally, each volume was normalized to have zero mean and unit variance, ensuring consistent voxel value ranges.
Preprocessing also included selecting representative 2D slices from each volume. From the center of each volume along the axial plane, we extracted four consecutive middle slices and repeated them across three channels, creating a 3-channel 2D image in an RGB-like format.
To balance the dataset, we augmented only the AD group by applying random rotations within a range of -5° to +5°. For the prediction phase, we utilized 390 sMCI subjects, who had not converted to AD, and 140 pMCI subjects, who had converted to AD within 48 months from baseline visit, to predict their status at month 48, using data from four consecutive time points at 6-month intervals: baseline, month 06, month 12, and month 18. A forward pass was used for missing time points (month 12 was used in place of month 18, when unavailable). Only the middle slice of volumes was selected for prediction. These time points were combined to form a sequence of four observations for each subject.
To enhance data balance and improve the robustness of prediction, we applied a random rotation of -5° to +5° to the pMCI subjects during the diagnosis phase in inference mode. After augmentation, the final dataset used in the prediction phase consisted of 390 sMCI subjects and 420 pMCI subjects.

\subsection{Implementation details}
We applied 5-fold cross-validation during training for both the feature extraction and prediction models. In training EffNetViTLoRA, four different ranks (4, 8, 16, and 32) were tested for the LoRA matrices. The best performance was achieved with rank 8, which was subsequently applied to all encoder blocks of the ViT for adapting the key, query, and value matrices. This configuration resulted in a highly efficient model with only 419K learnable parameters in EffNetViTLoRA feature extraction model, a significant reduction compared to training a full Vision Transformer from scratch. This highlights the efficiency and adaptability of the LoRA approach, making it a promising solution for applying Vision Transformers to medical imaging tasks, where data availability is often limited.
 
All experiments were conducted on an NVIDIA RTX 3090 GPU \cite{cuda}. The number of training epochs was set to 100 for the feature extraction phase and 400 for the prediction phase. Batch sizes were fixed at 32 for EffNetViTLoRA and 64 for the prediction model. We used the Adam optimizer \cite{kingma2014adam} with a learning rate of 1e-3 for the feature extraction model. In the prediction model, the learning rate was set to 1e-3 for the first 200 epochs to accelerate the learning process, and then reduced to 1e-4 for the final 200 epochs to allow the model to carefully tune the weights using a learning rate scheduler.
The prediction model consisted of a single BiLSTM layer with 256 units, to keep the number of trainable parameters reasonable. A dropout rate of 0.5 was applied to prevent overfitting. Cross-entropy loss was employed to train the EffNetViTLoRA model. The categorical cross-entropy loss for a sample with ground truth label $ {y} $ and predicted probability distribution $ \hat{y} $ is defined as:
\begin{equation}
	\mathcal{L}_{CE} = - \sum_{c=1}^{C} y_c \log(\hat{y}_c),
\end{equation}
where $ {C} $ is the number of classes, $ {y}_c $ is the one-hot encoded ground truth, and $ \hat{y}_c $ is the predicted probability for class c.

Given the challenging nature of the prediction task, distinguishing between sMCI and pMCI, where subtle differences exist, we further employed the Focal Loss function \cite{lin2017focal}. The Focal Loss extends cross-entropy by introducing a modulating factor that reduces the loss contribution from well-classified examples, enabling the model to focus on more difficult samples:
\begin{equation}
	\mathcal{L}_{FL} = - \sum_{c=1}^{C} \alpha_c \, y_c \, (1 - \hat{y}_c)^{\gamma} \, \log(\hat{y}_c),
\end{equation}
where $ \hat{y}_c $ is the predicted probability of the true class, $ \alpha $ is a balancing factor that set to 1 in our experiments, because we made our data balanced by augmentation, and $ \gamma $ is a focusing parameter that set to 2 in our experiments. By applying focal loss function, the model places greater emphasis on misclassified and ambiguous cases, which is particularly beneficial for the prediction model.

\subsection{Evaluation Metrics}
We evaluated the both our models performances using common metrics, including accuracy, precision, recall, and F1-score, which are mathematically defined as follows:
\begin{equation}
	Accuracy = \frac {TP + TN}{TP + TN + FP + FN}
\end{equation}
\begin{equation}
	Precision = \frac{TP}{TP + FP}
\end{equation}
\begin{equation}
	Recall = \frac{TP}{TP + FN}
\end{equation}
\begin{equation}
	\text{F1-score} = 2 \times \frac{Precision \times Recall}{Precision + Recall}
\end{equation}
Where TP, TN, FP, and FN stand for True Positive, True Negative, False Positive, and False Negative, respectively.
In medical research, recall is particularly important, as a high recall indicates that a large proportion of true positive cases are correctly identified. Similarly, high precision implies that a high percentage of identified positive cases are truly positive, meaning few healthy individuals are incorrectly classified as positive.

\subsection{Results}
By combining local and global features that capture dependencies within an image, the feature extractor model demonstrates promising performance in AD diagnosis \cite{khatooni2025effnetvitloraefficienthybriddeep}. This indicates that the extracted features are sufficiently representative to be employed in the subsequent prediction phase. Table \ref{tab:prediction_results} presents the prediction results obtained through 5-fold cross-validation, using temporal features to project the status 2.5 years beyond the last visit. Results from other studies on AD prediction are also included in the table for comparison. Our proposed two-stage method achieved 95.05\%, 93.98\%, 96.66\%, and 95.30\% in accuracy, precision, recall, and F1-score, respectively, outperforming existing state-of-the-art studies on AD prediction using follow-up data 2.5 years after the last visit. Although \cite{abuhmed2021robust} and \cite{el2022two} also performed prediction at month 48, we do not directly compare against them in the table, since their setting included AD, MCI, and CN groups together rather than focusing solely on MCI cases. We argue that transitions within the AD group have limited interpretive value, as AD patients are already fully affected by neurodegeneration. Moreover, direct transitions from CN to AD are not typically observed in the ADNI dataset. Thus, tracking CN subjects adds less value in this context. Instead, we emphasize that the most critical prediction task is distinguishing between MCI subjects who will convert to AD and those who will remain stable over time. Nevertheless, the referenced studies reported accuracies of 82.63\% and 93.87\% for AD vs. MCI vs. CN prediction, which indicates that our model outperforms such multi-class prediction approaches.

\begin{figure*}
	\centering
	\begin{subfigure}{0.45\textwidth}
		\centering
		\includegraphics[width=\textwidth]{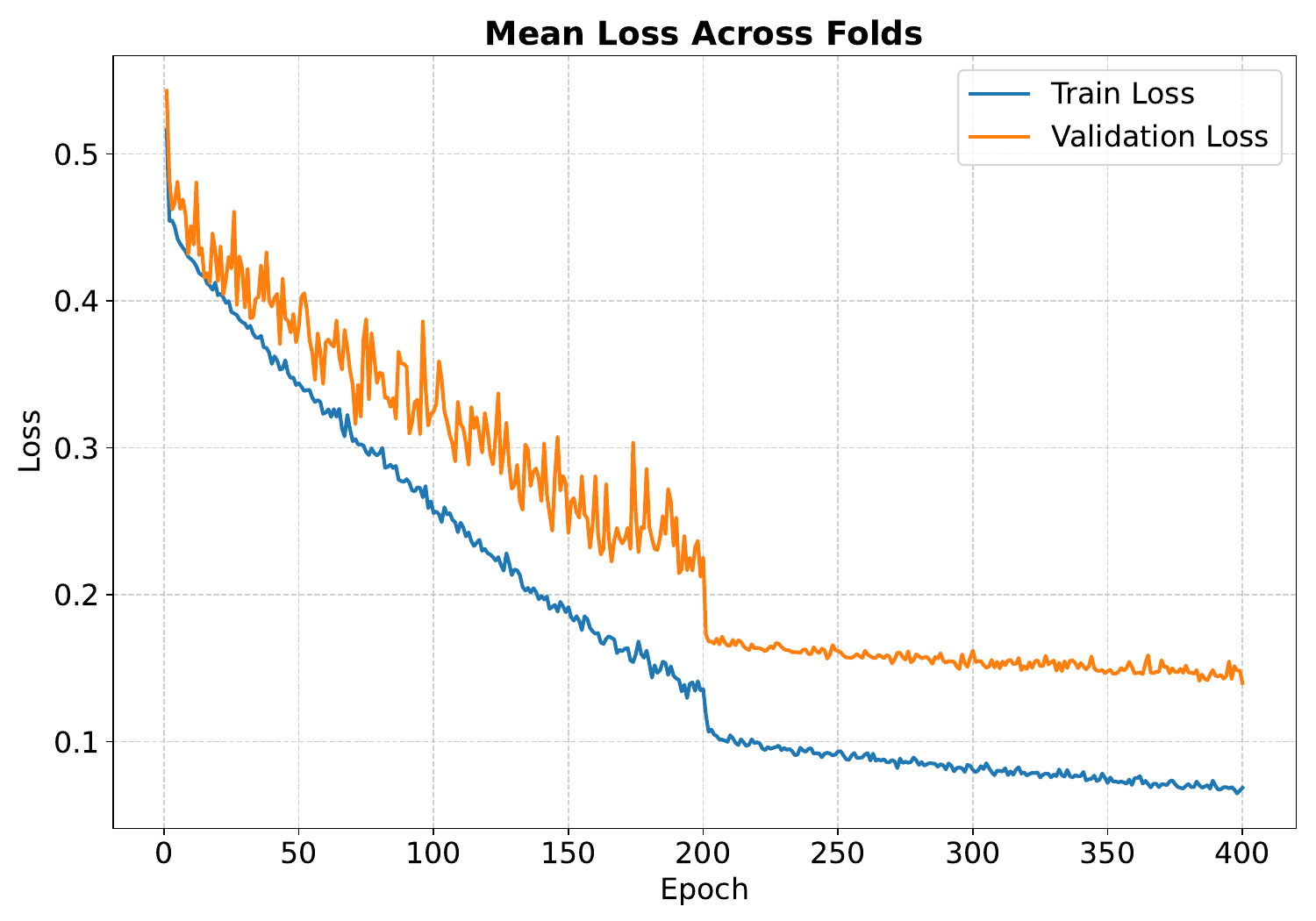}
		\caption{}
		\label{fig:predLoss}
	\end{subfigure}
	\begin{subfigure}{0.45\textwidth}
		\centering
		\includegraphics[width=\textwidth]{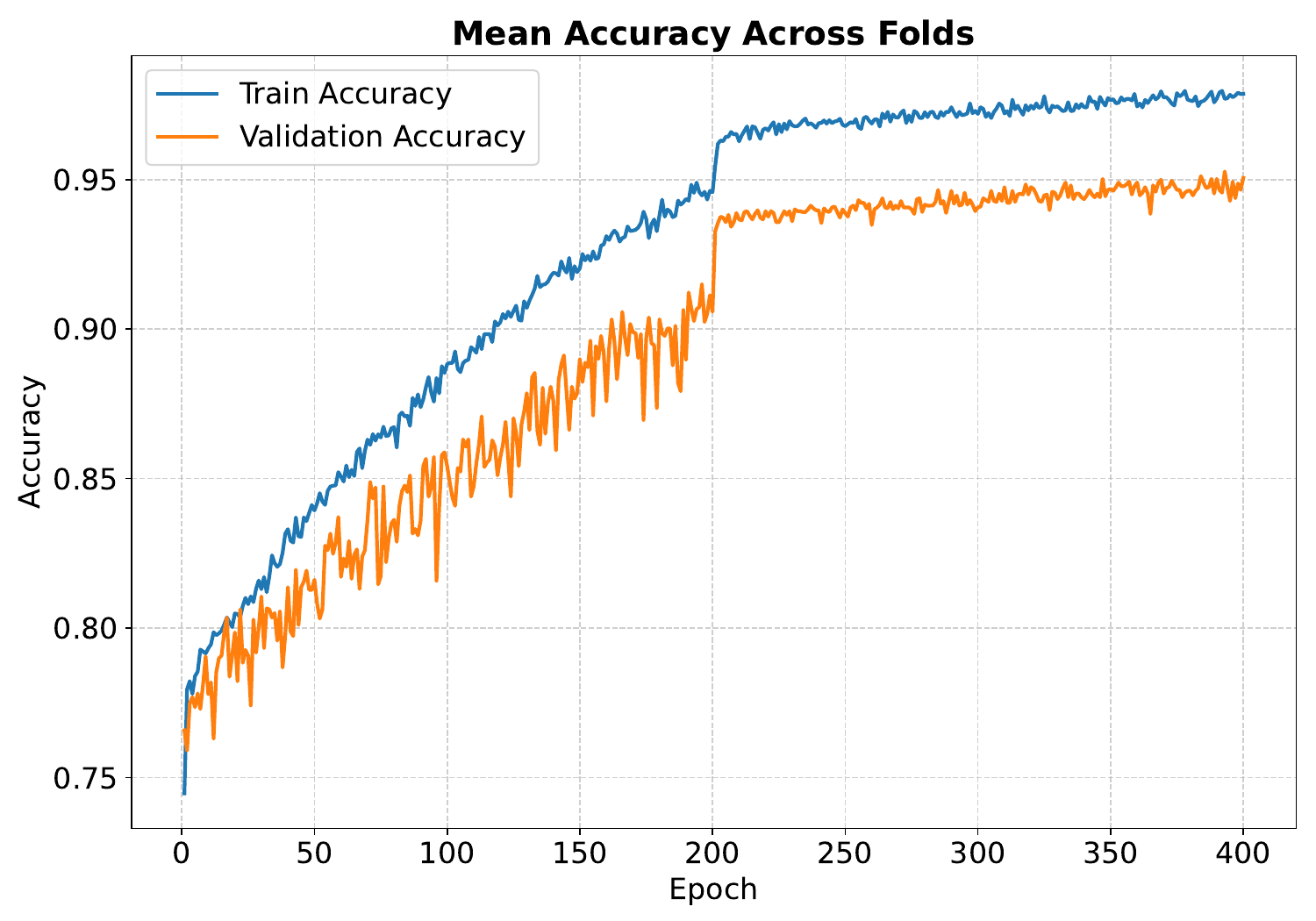}
		\caption{}
		\label{fig:predAcc}
	\end{subfigure}
	\caption{Plots of loss and accuracy. (a) shows the loss curve using the focal loss function, and (b) shows the accuracy curve during training. During the first 200 epochs, a higher learning rate was used to accelerate learning, and it was reduced in the following 200 epochs to allow for more precise optimization.}
	\label{fig:loss+acc}
\end{figure*}

\begin{figure*}
	\centering
	\includegraphics[scale=0.5]{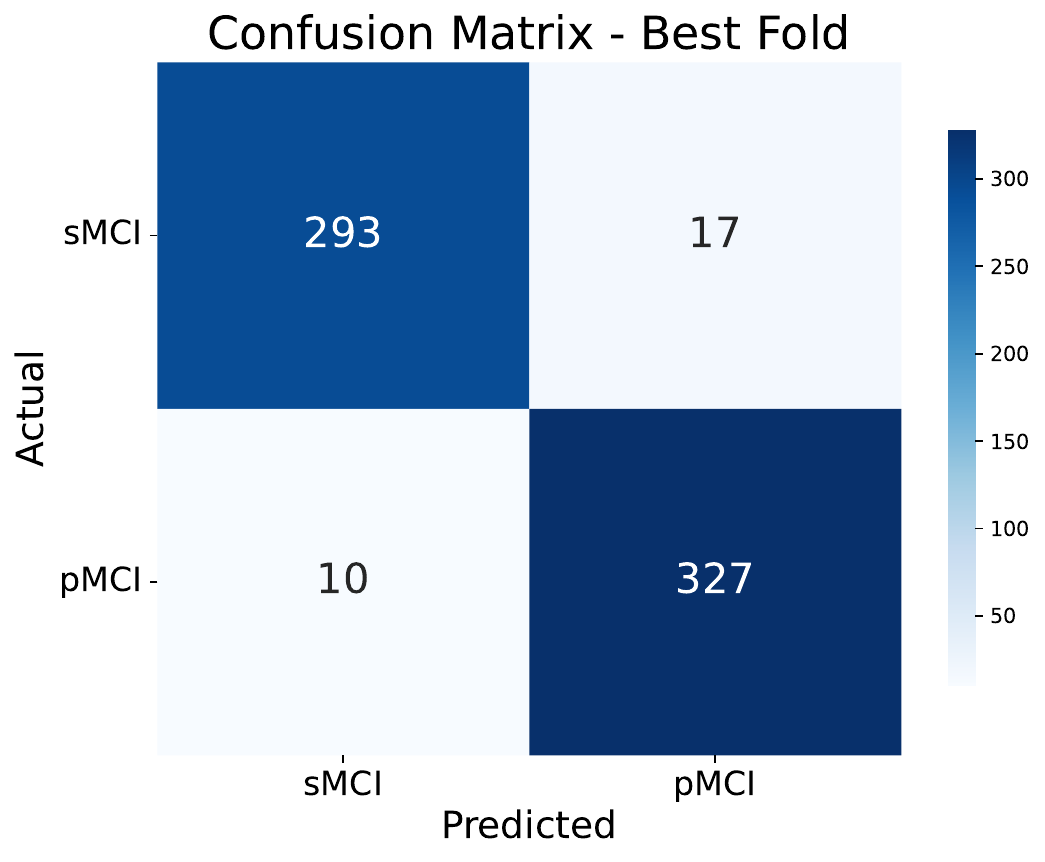}
	\caption{Confusion matrix of the prediction model: sMCI vs. pMCI prediction within 48 months, using the best-performing model from 5-fold cross-validation.}
	\label{fig:conf_matrix}
\end{figure*}

The temporal features consist of four vectorized features for each sample, which are a combination of image features extracted from the EffNetViTLoRA and 17 non-image biomarkers. We evaluated several groups of non-imaging biomarkers, including psychological clinical test values, demographic information such as age and sex, and engineered features from the ADNI dataset. By increasing the number of non-image biomarkers, we found that there was no improvement in the prediction results. Therefore, we aimed to keep the minimum number of effective biomarkers to maintain model efficiency during training and avoid overcomplicating the model. Finally, we selected 17 non-image neuropsychological biomarkers. The list of selected biomarkers includes: CDRSB, ADAS11, ADAS13, ADASQ4, MMSE, RAVLT\_immediate, RAVLT\_learning, RAVLT\_forgetting, RAVLT\_perc\_forgetting, FAQ, Ventricles, Hippocampus, WholeBrain, Entorhinal, Fusiform, MidTemp and ICV.
Our prediction model achieved an average training accuracy of 97.86\% and a validation accuracy of 95.05\% across five folds in predicting Alzheimer’s disease status within an 18-month follow-up period, projecting outcomes up to 2.5 years beyond the last visit. As Alzheimer’s disease is an irreversible neurodegenerative disorder, it is valuable to predict AD at the stage when primary symptoms first emerge in the brain. Distinguishing between stable MCI and progressive MCI cases is particularly challenging due to the subtle differences in brain features between the two groups. Nevertheless, the results highlight the model’s strong ability to separate these classes.
Figures \ref{fig:predLoss} and \ref{fig:predAcc} illustrate the training loss and accuracy curves respectively. The steadily decreasing loss curve indicates stable learning and effective convergence. To optimize training, we initially used a higher learning rate of 1e-3 for the first 200 epochs to accelerate convergence toward an optimal region of the parameter space. We then reduced the learning rate to 1e-4, allowing the model to fine-tune weights more effectively.
Because the dataset was balanced through augmentation, we set $ \alpha = 1 $ in the Focal Loss function. The average training loss reached approximately 6.86\%, while the validation loss stabilized around 13.99\% across five folds. To further examine model performance, we generated a confusion matrix, which is presented in Fig. \ref{fig:conf_matrix} from the best-performing fold. This analysis shows that, while the model generally distinguishes sMCI from pMCI effectively, misclassifications persist, reflecting the inherent difficulty of the task given the overlapping and subtle nature of the two groups’ features.
\begin{table*}
	\centering
	\caption{Comparison of performance metrics between the proposed two-stage prediction model and other studies on Alzheimer’s disease prediction. Our model predicts AD at month 48 using 5-fold cross-validation. Results from other studies are reported under similar conditions, using four time-point longitudinal data to predict disease progression within 48 months.}
	\label{tab:prediction_results}
	\renewcommand{\arraystretch}{1.2} 
	\setlength{\tabcolsep}{4pt} 
	\resizebox{\textwidth}{!}{ 
		\begin{tabular}{@{}lp{2.5cm}p{2cm}cccccc@{}}
			\toprule
			\textbf{Method}      & \textbf{Dataset} & \textbf{\# of Subjects (sMCI+pMCI)} & \textbf{Multimodality} & \textbf{Accuracy (\%)} & \textbf{Precision (\%)} & \textbf{Recall (\%)} & \textbf{F1 Score (\%)} \\ \midrule
			Our prediction model & ADNI    & 530                   & $\checkmark$             & \textbf{95.05} & \textbf{93.98} & \textbf{96.66} & \textbf{95.30}        \\
			\cite{zhu2021long}                      & ADNI    & 151                   & $\checkmark$                 & 85.4                   & -                    & -                 & -                   \\
			\cite{er2020predicting}                      & ADNI    & 126                   & $\times$                 & 87.2                   & 80.4                    & 92.4                & -                   \\
			\cite{hoang2023vision}                      & ADNI    & 598                   & $\times$                 & 83.27                   & 85.07                    & 81.48                & -                   \\
			\cite{hu2023vgg}                      & ADNI    & 275                   & $\times$                 & 77.20                   & 71.59                    & 79.97                & -                   
			\\
			\bottomrule
	\end{tabular}}
\end{table*}
\subsection{Ablation Studies}
\label{sec:ablation}
\subsubsection{Input sequence without non-imaging (neuropsychological) biomarkers}
As mentioned in Section \ref{sec:prediction_model}, we constructed feature sequences at each time point (baseline, month 6, month 12, and month 18), where each feature vector comprised 256 imaging features and 17 non-imaging biomarkers. To evaluate the contribution of non-imaging biomarkers in our proposed method, we conducted an ablation experiment by removing them and using only the 256-dimensional imaging features extracted from the feature extractor model. Table \ref{tab:nonimage_ablation} presents the results in comparison with our full proposed approach. The findings indicate that incorporating non-imaging biomarkers, such as whole brain volume, hippocampus volume and other clinical measures, substantially improves predictive performance. This demonstrates the importance of integrating imaging and non-imaging features to capture complementary information for Alzheimer’s disease prediction.
\begin{table}
	\centering
	\caption{Ablation study on the contribution of non-imaging biomarkers. The best results are highlighted in bold.}
	\label{tab:nonimage_ablation}
	\renewcommand{\arraystretch}{1.2}
	\resizebox{\textwidth}{!}{
		\begin{tabular}{lcccc}
			\hline
			\textbf{Features} & \textbf{Accuracy (\%)} & \textbf{Precision (\%)} & \textbf{Recall (\%)} & \textbf{F1-score (\%)} \\
			\hline
			MRI image features only & 71.47 & 73.97 & 69.81 & 71.74 \\
			MRI features + Non-imaging features (proposed) & \textbf{95.05} & \textbf{93.98} & \textbf{96.66} & \textbf{95.30} \\
			\hline
	\end{tabular}}
\end{table}
\subsubsection{Vanilla LSTM}
In a bidirectional LSTM, the model processes sequences in both forward and backward directions, enabling it to capture contextual information from the entire sequence. In contrast, a vanilla LSTM only captures dependencies in the forward temporal order. BiLSTM is particularly effective when all time points are available simultaneously, though it comes at the cost of doubling the number of trainable parameters. To evaluate this trade-off, we also tested the vanilla LSTM to examine whether only forward analysis could effectively model the relationships between feature tokens. Table~\ref{tab:lstm_ablation} reports the performance metrics and the number of trainable parameters. The BiLSTM consistently outperforms the vanilla LSTM across all metrics by a large difference, highlighting the benefit of leveraging temporal information in both forward and backward directions for richer feature representation.
\begin{table}
	\centering
	\caption{Ablation study comparing vanilla LSTM and BiLSTM models on the validation set using 5-fold cross-validation. The best results are highlighted in bold.}
	\label{tab:lstm_ablation}
	\renewcommand{\arraystretch}{1.2}
	\resizebox{\textwidth}{!}{
		\begin{tabular}{lccccc}
			\hline
			\textbf{Model} & \textbf{Accuracy (\%)} & \textbf{Precision (\%)} & \textbf{Recall (\%)} & \textbf{F1-score (\%)} & \textbf{Parameters (M)} \\
			\hline
			Vanilla LSTM   & 84.23 & 83.29 & 87.37 & 85.18 & 0.5 \\
			Bidirectional LSTM (proposed) & \textbf{95.05} & \textbf{93.98} & \textbf{96.66} & \textbf{95.30} & 1 \\
			\hline
	\end{tabular}}
\end{table}
\subsubsection{Loss function}
To evaluate the impact of different loss functions on model performance, we conducted an ablation study by replacing the focal loss with the binary cross-entropy (BCE) loss. While BCE is widely used for binary classification tasks, focal loss is specifically designed to address class imbalance by down-weighting easy examples and emphasizing harder ones. Table \ref{tab:loss_ablation} presents the obtained metrics using both BCE and focal loss functions. As shown in the table, focal loss outperforms the standard BCE loss, achieving higher accuracy, precision, recall, and F1-score. This suggests that focal loss is more effective in handling the challenging nature of our dataset by focusing on hard samples.
\begin{table}
	\centering
	\caption{Ablation study on different loss functions. The best results are highlighted in bold.}
	\label{tab:loss_ablation}
	\renewcommand{\arraystretch}{1.2}
	\resizebox{\textwidth}{!}{
	\begin{tabular}{lcccc}
		\hline
		\textbf{Loss Function} & \textbf{Accuracy (\%)} & \textbf{Precision (\%)} & \textbf{Recall (\%)} & \textbf{F1-score (\%)} \\
		\hline
		Binary Cross-Entropy (BCE) & 92.95 & 91.52 & 95.25 & 93.34 \\
		Focal Loss (proposed)      & \textbf{95.05} & \textbf{93.98} & \textbf{96.66} & \textbf{95.30} \\
		\hline
	\end{tabular}}
\end{table}
\section{Conclusion}
\label{sec:conclusion}
In this paper, we proposed a two-phase multimodal model for Alzheimer’s disease prediction. In the first phase, we employed EffNetViTLoRA as the feature extractor, using the MRI modality. We utilized all available MRI data from the ADNI cohorts (ADNI1, ADNIGO, ADNI2, and ADNI3), ensuring a dataset free from selection bias and enhancing model generalizability. EffNetViTLoRA is a hybrid architecture that combines EfficientNet with ViTLoRA. ViTLoRA adapts the Vision Transformer pretrained on ImageNet-1K using LoRA, with low-rank decomposable matrices enabling the model to learn differences between the source and target datasets without altering the pretrained weights. In the second phase, the extracted MRI features from the first phase, along with additional non-image biomarkers, were used to construct a longitudinal dataset across four time points: baseline, month 06, month 12, and month 18. This follow-up dataset was then used in the prediction phase, which consisted of a BiLSTM layer with 256 hidden units to process and analyze relationships between sequences of input tokens in both forward and backward directions, and a single output neuron. The proposed model achieved an average accuracy of 95.05\% across five folds in predicting stable MCI versus progressive MCI at month 48, outperforming comparable studies on sMCI vs. pMCI prediction. These results highlight its potential as a reliable and effective approach for early Alzheimer's disease prediction.

Future work will focus on incorporating specific brain regions known to be affected by Alzheimer’s disease, such as segmented gray matter, segmented white matter, and hippocampi, instead of full MRI volumes. This approach aims to help the model concentrate on the most relevant areas, thereby accelerating learning, improving prediction performance, and potentially reducing misdiagnosis rates.

The code for this project is available at \url{https://github.com/MBehjat/AD-Prediction}.

  	\bibliographystyle{elsarticle-num} 
	\bibliography{library}

@article{patterson2018world,
  title={World Alzheimer Report 2018},
  author={Patterson, C},
  journal={The State of the Art of Dementia Research: New Frontiers},
  year={2018}
}

@article{el2022two,
  title={Two-stage deep learning model for Alzheimer’s disease detection and prediction of the mild cognitive impairment time},
  author={El-Sappagh, Shaker and Saleh, Hager and Ali, Farman and Amer, Eslam and Abuhmed, Tamer},
  journal={Neural Computing and Applications},
  volume={34},
  number={17},
  pages={14487--14509},
  year={2022},
  publisher={Springer}
}

@article{el2020multimodal,
  title={Multimodal multitask deep learning model for Alzheimer’s disease progression detection based on time series data},
  author={El-Sappagh, Shaker and Abuhmed, Tamer and Islam, SM Riazul and Kwak, Kyung Sup},
  journal={Neurocomputing},
  volume={412},
  pages={197--215},
  year={2020},
  publisher={Elsevier}
}

@article{lei2020deep,
  title={Deep and joint learning of longitudinal data for Alzheimer's disease prediction},
  author={Lei, Baiying and Yang, Mengya and Yang, Peng and Zhou, Feng and Hou, Wen and Zou, Wenbin and Li, Xia and Wang, Tianfu and Xiao, Xiaohua and Wang, Shuqiang},
  journal={Pattern Recognition},
  volume={102},
  pages={107247},
  year={2020},
  publisher={Elsevier}
}

@article{simonyan2014very,
  title={Very deep convolutional networks for large-scale image recognition},
  author={Simonyan, Karen},
  journal={arXiv preprint arXiv:1409.1556},
  year={2014}
}

@article{lecun2015deep,
  title={Deep learning},
  author={LeCun, Yann and Bengio, Yoshua and Hinton, Geoffrey},
  journal={nature},
  volume={521},
  number={7553},
  pages={436--444},
  year={2015},
  publisher={Nature Publishing Group UK London}
}

@article{dosovitskiy2020image,
  title={An image is worth 16x16 words: Transformers for image recognition at scale},
  author={Dosovitskiy, Alexey},
  journal={arXiv preprint arXiv:2010.11929},
  year={2020}
}

@article{vaswani2017attention,
  title={Attention is all you need},
  author={Vaswani, A},
  journal={Advances in Neural Information Processing Systems},
  year={2017}
}

@article{cavedo2014medial,
  title={Medial temporal atrophy in early and late-onset Alzheimer's disease},
  author={Cavedo, Enrica and Pievani, Michela and Boccardi, Marina and Galluzzi, Samantha and Bocchetta, Martina and Bonetti, Matteo and Thompson, Paul M and Frisoni, Giovanni B},
  journal={Neurobiology of aging},
  volume={35},
  number={9},
  pages={2004--2012},
  year={2014},
  publisher={Elsevier}
}

@article{tustison2010n4itk,
  title={N4ITK: improved N3 bias correction},
  author={Tustison, Nicholas J and Avants, Brian B and Cook, Philip A and Zheng, Yuanjie and Egan, Alexander and Yushkevich, Paul A and Gee, James C},
  journal={IEEE transactions on medical imaging},
  volume={29},
  number={6},
  pages={1310--1320},
  year={2010},
  publisher={IEEE}
}

@article{isensee2019automated,
  title={Automated brain extraction of multisequence MRI using artificial neural networks},
  author={Isensee, Fabian and Schell, Marianne and Pflueger, Irada and Brugnara, Gianluca and Bonekamp, David and Neuberger, Ulf and Wick, Antje and Schlemmer, Heinz-Peter and Heiland, Sabine and Wick, Wolfgang and others},
  journal={Human brain mapping},
  volume={40},
  number={17},
  pages={4952--4964},
  year={2019},
  publisher={Wiley Online Library}
}

@article{woolrich2009bayesian,
  title={Bayesian analysis of neuroimaging data in FSL},
  author={Woolrich, Mark W and Jbabdi, Saad and Patenaude, Brian and Chappell, Michael and Makni, Salima and Behrens, Timothy and Beckmann, Christian and Jenkinson, Mark and Smith, Stephen M},
  journal={Neuroimage},
  volume={45},
  number={1},
  pages={S173--S186},
  year={2009},
  publisher={Elsevier}
}

@article{jenkinson2001global,
  title={A global optimisation method for robust affine registration of brain images},
  author={Jenkinson, Mark and Smith, Stephen},
  journal={Medical image analysis},
  volume={5},
  number={2},
  pages={143--156},
  year={2001},
  publisher={Elsevier}
}

@article{jenkinson2002improved,
  title={Improved optimization for the robust and accurate linear registration and motion correction of brain images},
  author={Jenkinson, Mark and Bannister, Peter and Brady, Michael and Smith, Stephen},
  journal={Neuroimage},
  volume={17},
  number={2},
  pages={825--841},
  year={2002},
  publisher={Elsevier}
}

@inproceedings{deng2009imagenet,
  title={Imagenet: A large-scale hierarchical image database},
  author={Deng, Jia and Dong, Wei and Socher, Richard and Li, Li-Jia and Li, Kai and Fei-Fei, Li},
  booktitle={2009 IEEE conference on computer vision and pattern recognition},
  pages={248--255},
  year={2009},
  organization={Ieee}
}

@inproceedings{tan2021efficientnetv2,
  title={Efficientnetv2: Smaller models and faster training},
  author={Tan, Mingxing and Le, Quoc},
  booktitle={International conference on machine learning},
  pages={10096--10106},
  year={2021},
  organization={PMLR}
}

@article{hu2021lora,
  title={Lora: Low-rank adaptation of large language models},
  author={Hu, Edward J and Shen, Yelong and Wallis, Phillip and Allen-Zhu, Zeyuan and Li, Yuanzhi and Wang, Shean and Wang, Lu and Chen, Weizhu},
  journal={arXiv preprint arXiv:2106.09685},
  year={2021}
}

@article{kingma2014adam,
  title={Adam: A method for stochastic optimization},
  author={Kingma, Diederik P},
  journal={arXiv preprint arXiv:1412.6980},
  year={2014}
}

@misc{AlzheimerSymptoms,
  author = {hopkinsmedicine},
  title  = {Stages of Alzheimer's Disease},
  year   = {2024},
  url    = {https://www.hopkinsmedicine.org/health/conditions-and-diseases/alzheimers-disease/stages-of-alzheimer-disease},
  note   = {Accessed: February 2, 2025}
}

@misc{cuda,
  author={NVIDIA and Vingelmann, Péter and Fitzek, Frank H.P.},
  title={CUDA, release: 10.2.89},
  year={2020},
  url={https://developer.nvidia.com/cuda-toolkit},
}

@inproceedings{lin2017focal,
  title={Focal loss for dense object detection},
  author={Lin, Tsung-Yi and Goyal, Priya and Girshick, Ross and He, Kaiming and Doll{\'a}r, Piotr},
  booktitle={Proceedings of the IEEE international conference on computer vision},
  pages={2980--2988},
  year={2017}
}

@article{zhu2021long,
  title={Long range early diagnosis of Alzheimer's disease using longitudinal MR imaging data},
  author={Zhu, Yingying and Kim, Minjeong and Zhu, Xiaofeng and Kaufer, Daniel and Wu, Guorong and Alzheimer's Disease Neuroimaging Initiative and others},
  journal={Medical image analysis},
  volume={67},
  pages={101825},
  year={2021},
  publisher={Elsevier}
}

@article{er2020predicting,
  title={Predicting the prognosis of MCI patients using longitudinal MRI data},
  author={Er, Fusun and Goularas, Dionysis},
  journal={IEEE/ACM Transactions on Computational Biology and Bioinformatics},
  volume={18},
  number={3},
  pages={1164--1173},
  year={2020},
  publisher={IEEE}
}

@article{abuhmed2021robust,
  title={Robust hybrid deep learning models for Alzheimer’s progression detection},
  author={Abuhmed, Tamer and El-Sappagh, Shaker and Alonso, Jose M},
  journal={Knowledge-Based Systems},
  volume={213},
  pages={106688},
  year={2021},
  publisher={Elsevier}
}

@article{hoang2023vision,
  title={Vision transformers for the prediction of mild cognitive impairment to Alzheimer’s disease progression using mid-sagittal sMRI},
  author={Hoang, Gia Minh and Kim, Ue-Hwan and Kim, Jae Gwan},
  journal={Frontiers in Aging Neuroscience},
  volume={15},
  pages={1102869},
  year={2023},
  publisher={Frontiers Media SA}
}

@article{hu2023vgg,
  title={VGG-TSwinformer: Transformer-based deep learning model for early Alzheimer’s disease prediction},
  author={Hu, Zhentao and Wang, Zheng and Jin, Yong and Hou, Wei},
  journal={Computer Methods and Programs in Biomedicine},
  volume={229},
  pages={107291},
  year={2023},
  publisher={Elsevier}
}

@article{gao2023hybrid,
  title={A hybrid multi-scale attention convolution and aging transformer network for Alzheimer's disease diagnosis},
  author={Gao, Xingyu and Cai, Hongjie and Liu, Manhua},
  journal={IEEE Journal of Biomedical and Health Informatics},
  volume={27},
  number={7},
  pages={3292--3301},
  year={2023},
  publisher={IEEE}
}

@misc{khatooni2025effnetvitloraefficienthybriddeep,
      title={EffNetViTLoRA: An Efficient Hybrid Deep Learning Approach for Alzheimer's Disease Diagnosis}, 
      author={Mahdieh Behjat Khatooni and Mohsen Soryani},
      year={2025},
      eprint={2508.19349},
      archivePrefix={arXiv},
      primaryClass={cs.CV},
      url={https://arxiv.org/abs/2508.19349}, 
}

@article{borbely2013neuropeptides,
  title={Neuropeptides in learning and memory},
  author={Borb{\'e}ly, {\'E}va and Scheich, B{\'a}lint and Helyes, Zsuzsanna},
  journal={Neuropeptides},
  volume={47},
  number={6},
  pages={439--450},
  year={2013},
  publisher={Elsevier}
}
\end{document}